\documentclass{article}
\usepackage{spconf,amsmath,graphicx}
\usepackage{bbding}
\usepackage{setspace}

\graphicspath{{figures/}}

\begin{document}\sloppy

\title{Boundary Corrected Multi-scale Fusion Network for Real-time Semantic Segmentation}

\name{Tianjiao Jiang, Yi Jin*, Tengfei Liang, Xu Wang, Yidong Li \thanks{*Corresponding author.} \vspace{-8pt}}
\address{
    School of Computer and Information Technology, Beijing Jiaotong University, Beijing, China\\
  	\{tianjiao.jiang, yjin, tengfei.liang, xu.wang, ydli\}@bjtu.edu.cn
    \vspace{-8pt}
}

\maketitle

\begin{abstract}
Image semantic segmentation aims at the pixel-level classification of images, which has requirements for both accuracy and speed in practical application.
Existing semantic segmentation methods mainly rely on the high-resolution input to achieve high accuracy and do not meet the requirements of inference time.
Although some methods focus on high-speed scene parsing with lightweight architectures, they can not fully mine semantic features under low computation with relatively low performance.
To realize the real-time and high-precision segmentation, we propose a new method named Boundary Corrected Multi-scale Fusion Network, which uses the designed Low-resolution Multi-scale Fusion Module to extract semantic information.
Moreover, to deal with boundary errors caused by low-resolution feature map fusion, we further design an additional Boundary Corrected Loss to constrain overly smooth features.
Extensive experiments show that our method achieves a state-of-the-art balance of accuracy and speed for the real-time semantic segmentation.
\end{abstract}


\section{Introduction}
As a basic vision task, image semantic segmentation \cite{Survey} is crucial for scene understanding. 
Its goal is to assign a semantic category label for each image pixel. 
With the development of deep learning and improved computing resources, Convolutional Neural Networks (CNN) are applied to image segmentation and significantly outperform traditional methods based on hand-crafted features. 
The end-to-end fully convolutional neural network \cite{FCN} method greatly promotes the rapid development of CNN in semantic segmentation. 
Then various forms of feature extraction and fusion modules have been proposed to improve the accuracy of the model \cite{Segnet,DeepLab,DeepLabv2, DeepLabv3}. 
However, most of the existing methods are designed to get high classification accuracy on each pixel with high-resolution images/feature maps and can not meet the speed requirements in deployment.
Therefore, some researchers recently focus on the designed of real-time and efficient models \cite{ENet,BiSeNet,BiSeNetv2}, which has more potential value in practical application.

\begin{figure}[t]
\centerline{\includegraphics[width=0.5\textwidth]{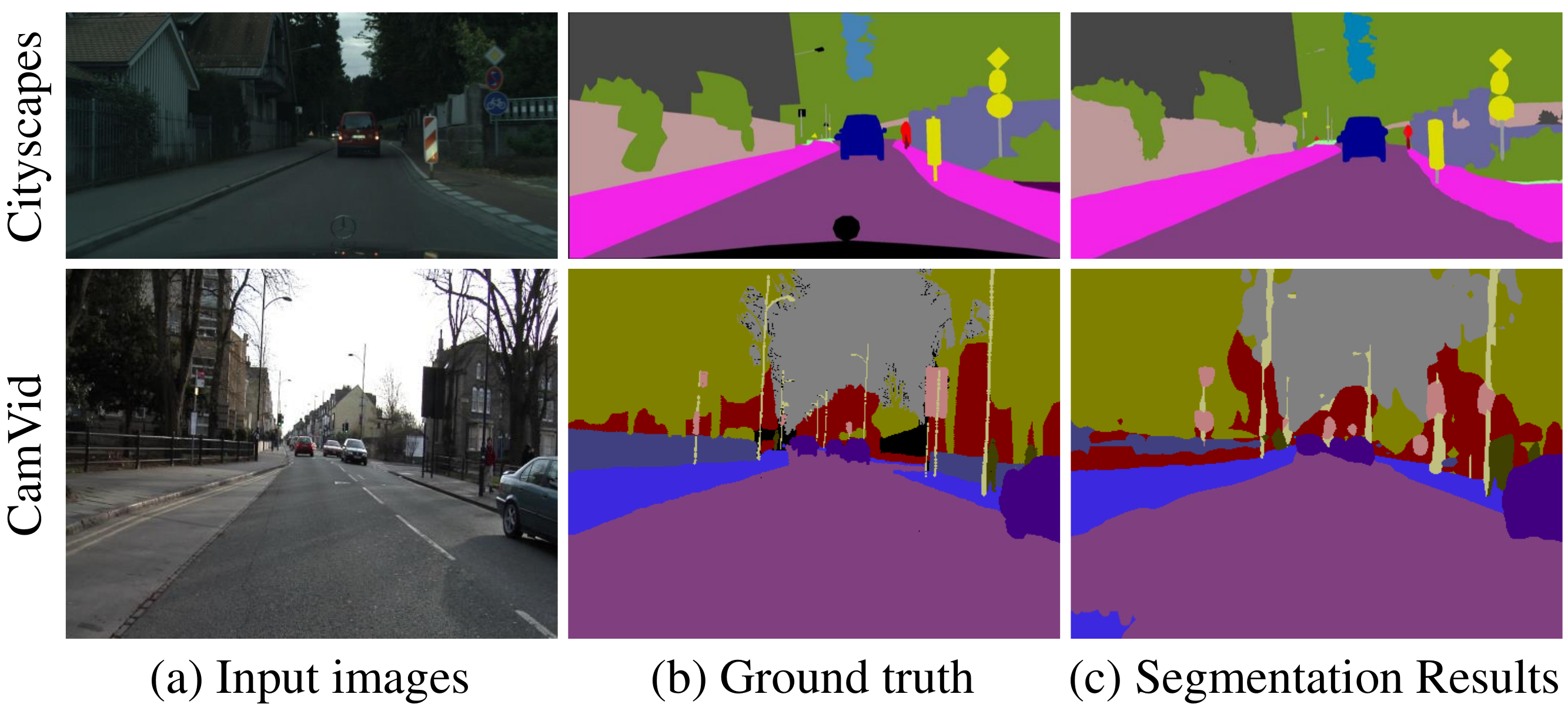}}
\vspace{-5pt}
\caption{Segmentation result of our proposed BCMFNet on the test set of Cityscapes and CamVid datasets.}
\vspace{-16pt}
\label{results}
\end{figure}

Recently, several networks emerged based on various efficient backbone networks, such as \cite{res1, res2, res3} based on ResNet-18 \cite{res}. However, with the proposal of better backbone networks, these structures designed for specific backbones are difficult to migrate.
The rest start developing new lightweight networks.
BiSeNet\cite{BiSeNet} proposes a new dual-branch network to solve the problem of the limited receptive field.
ICNet\cite{ICNet} proposes cascaded networks that fuse the details of high- and low-resolution feature maps.
But their processing of high-resolution feature maps limits the speed of the network.
CABiNet\cite{CABiNet} proposes a dual-branch structure to extract spatial details and contextual information. DDRNet\cite{DDRNet} proposes a dual-resolution network and a cascaded multi-scale feature extraction module.
However, none of their processing for dual branches takes into account the fine boundary features of the image.

Based on the above observations, we propose a new method, the Boundary Corrected Multi-scale Fusion Network (BCMFNet), with multi-scale feature fusion and boundary corrected loss.
Considering the computational constraints of lightweight models, we propose a feature fusion method to perform multi-scale feature extraction on low-resolution feature maps.
It is used to eliminate the high computational cost caused by high-resolution feature maps calculation and obtain more contextual information.
In addition, for the problem of missing fine boundaries caused by the fusion of low-resolution feature maps, we use a boundary corrected loss to extract hard samples of boundaries, capture the long-distance information of feature maps, and improve the boundary perception of the model (Figure \ref{results}).

\begin{figure*}[t]
\centerline{\includegraphics[width=1\textwidth]{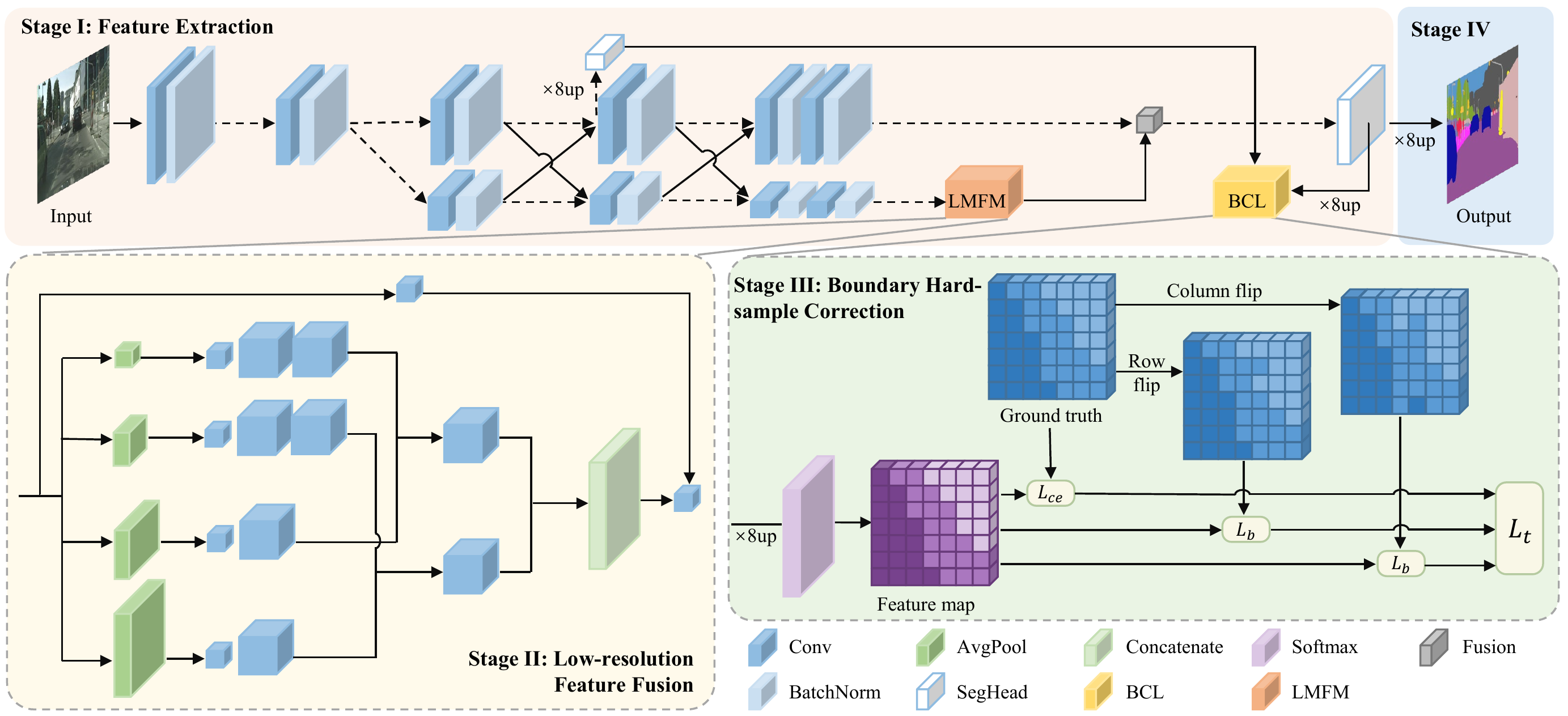}}
\vspace{-7pt}
\caption{
  The overview of our proposed BCMFNet method.
  The LMFM is marked in light yellow region, and BCL is marked in green region.
  Black solid lines denote information paths with data processing and black dashed lines denote information paths without data processing.}
\vspace{-13pt}
\label{sum}
\end{figure*}

Our contribution mainly includes four aspects:
(1) A new real-time semantic segmentation method is proposed, named Boundary Corrected Multi-scale Fusion Network (BCMFNet).
(2) A novel Low-resolution Multi-scale Fusion Module (LMFM) is designed to fuse low-resolution feature maps with multi-scale features to obtain richer contextual information.
(3) An additional Boundary Corrected Loss (BCL) function is introduced to enhance the learning of hard samples with correction of fine boundaries.
(4) Our method achieves a good balance between accuracy and speed with 78.2\% mIoU at 102 FPS on the cityscape dataset and 76.2\% mIoU at 230 FPS on the CamVid dataset.

\vspace{-8pt}
\section{Method}
\label{method}
\vspace{-5pt}
In this section, we introduce our boundary corrected multi-scale fusion network (BCMFNet) in detail. In Section \ref{overall}, we first describe the overall architecture of BCMFNet, including the network structure and the overall objective function during training. Then we introduce our proposed LMFM and BCL in the next two subsections.
\vspace{-6pt}
\subsection{Overall Structure}
\label{overall}
\vspace{-3pt}
As shown in Figure \ref{sum}, our method builds a novel network. The overall process of BCMFNet mainly includes four stages: feature extraction, low-resolution feature fusion, boundary hard sample correction, and upsampling.\par
At the feature extraction stage, we use DDRNet-s as the backbone network to fully fuse spatial and semantic information through multiple high- and low-resolution bilateral feature fusion.
At the low-resolution feature fusion stage, we use LMFM to extract the information of low-resolution feature maps. At the boundary hard sample correction stage, we use BCL to correct the boundary loss problem caused by over-smoothing. In the training mode, the overall objective function of BCMFNet is as follows:\par
{\setlength\abovedisplayskip{-7pt}
\setlength\belowdisplayskip{5pt}
\begin{align}
L{t} = L_{ce}(y_{pred}, y_{gt}) + \alpha L_{b}(y_{pred}, y_{gt}^{'} ) ,
\end{align}}
\vspace{-10pt}

\noindent where $L_{ce}$ is the cross-entropy loss, $L_{b}$ is the boundary loss, and $\alpha$ is a hyper-parameter to balance these two components. $y_{pred}$ is the prediction result obtained by softmax from the feature map. $y_{gt}$ and $y_{gt}^{'}$ are the ground truth before and after flipping. Finally, we use channel compression on the low-resolution feature maps using 1×1 convolutions and upsample them using bilinear interpolation.

\vspace{-6pt}
\subsection{Low-resolution Multi-scale Fusion Module}
\begin{figure}[t]
\centerline{\includegraphics[width=0.5\textwidth]{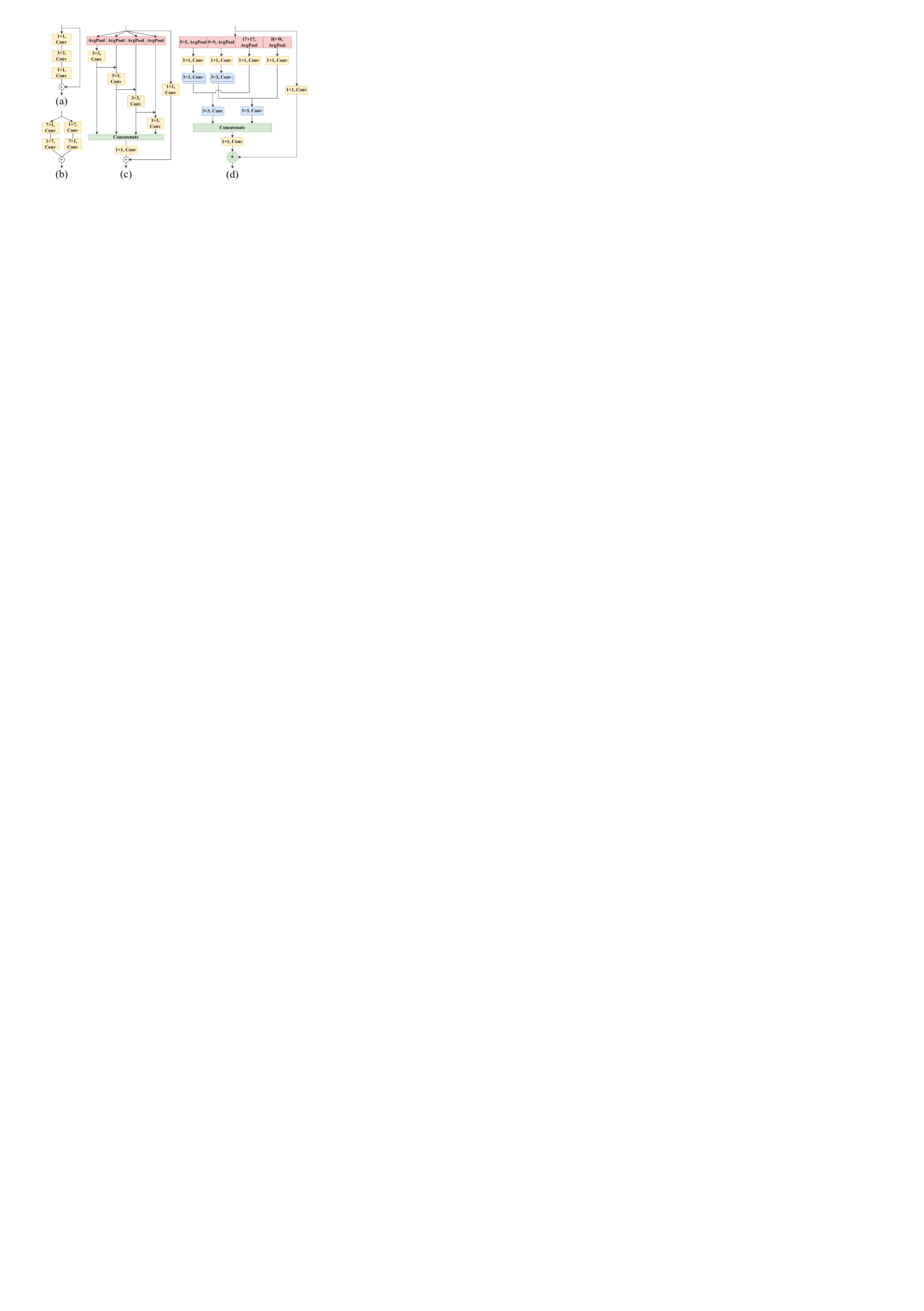}}
\caption{Comparison of different convolution blocks. (a) is bottleneck\cite{res}, (b) is GCN\cite{GCN}, (c) is DAPPM\cite{DDRNet}, and (d) is our proposed LMFM.}
\vspace{-15pt}
\label{compare}
\end{figure}

\vspace{-3pt}
The images contain objects of different sizes, and it is effective to process the feature maps at different scales proposed in global convolutional network (GCN) \cite{GCN} and Inception \cite{Inception}. As shown in Figure \ref{compare}, we compare different methods of extracting feature maps. The traditional bottleneck module \cite{res} uses 3×3 convolution kernels in each layer of the network, which limits the receptive field due to the small and fixed size of the convolution kernel. GCN adopts a multi-branch computational structure to improve the accuracy of segmentation, but a large amount of computation on high-resolution images limits the speed of the model. DDRNet-s \cite{DDRNet} uses a cascaded method to fuse the information of each layer upward, but the smooth information of the low layer has a great influence on the feature map of the high layer, and it is easy to lose the fine boundary information.\par
To address these issues, we propose a new module to extract contextual information from the low-resolution feature maps of the model. Figure \ref{compare}(d) shows the specific structure of the LMFM. After the feature map is fed into the module, the average pooling layer is used to mine more information, and gradually larger pooling kernels are used for generating feature maps with input image resolutions of 1/2, 1/4, and 1/8. To utilize the information generated by GAP, we perform feature fusion by using 3×3 and 1×1 convolution kernels multiple times. The fusion method adopts interval connection as shown in Figure \ref{compare}(d), and concatenates the generated feature maps. Compared with the connection method of Res2Net, this method can extract context information better. In addition, inspired by the design of the residual network, we also add a 1×1 convolution kernel for fast connection to prevent the loss of shallow information.\par
Inside the LMFM, the feature map extracted by the smaller pooling kernel is processed at a deeper level to obtain deeper information, combined with the shallower information received by the larger pooling kernel. Compared with the connection method of DAPPM, LMFM can combine information more effectively to form multi-scale attributes at different levels. Adding this module to the low-resolution feature map can obtain richer contextual information without affecting the inference speed of the model.\par

\vspace{-6pt}
\subsection{Boundary Corrected Loss}
\vspace{-3pt}

Semantic segmentation aims to obtain labels for all target image pixels. We believe that boundary pixels are more likely to generate hard samples for semantic segmentation. The model's learning of boundaries can be enhanced by detecting hard samples. The current steps of online hard sample processing are as follows: 1) get feature maps generated by CNN; 2) calculate loss between obtained feature maps and ground truth; 3) evaluate the reliability of samples by calculated loss; 4) select samples with the larger loss as hard samples. However,  this approach ignores the influence of boundaries on feature maps. In addition, boundary smoothing caused by multi-scale feature fusion is widespread.\par
Aiming at these problems, we propose a simple and effective flip boundary corrected loss function. This function captures long-distance relationships in isolated regions by flipping entire rows or columns of pixels and detects boundary-hard samples online. This method can correct the boundary smoothing caused by low-resolution feature fusion. We believe that the higher the loss, the greater the probability that the position is a boundary-hard sample. Specifically, the algorithm flow is shown in Stage III Figure \ref{sum}.\par
The process of BCL is as follows: 1) Use the softmax layer to process the feature map; 2) Row flip and column flip ground truth: flip adjacent pixels according to the set step size; 3) Calculate the cross-entropy loss between the original feature map and the original ground truth. Calculate the cross-entropy loss of the feature map and the ground truth transformed by the row and column; 4) Use the Non-Maximum Suppression to filter the obtained loss, arrange the loss from large to small, and select the boundary-hard samples according to the input threshold; 5) Calculate the weighted sum of the selected sample loss and the original cross-entropy loss. BCL can be expressed as:\par
{\setlength\abovedisplayskip{-5pt}
\setlength\belowdisplayskip{5pt}
\begin{align}
L_{b}(y_{pred}, y_{gt}^{'})&=-\frac{\lambda_{1}}{N}\sum_{i}\sum_{c=1}^{M}y_{row}log\left(y_{pred}\right)
\\& + \left[ -\frac{\lambda_{2}}{N}\sum_{i}\sum_{c=1}^{M}y_{col}log\left(y_{pred}\right)\right]\notag ,
\end{align}}

\noindent where $y_{pred}$ represents the predicted result of the pixel, $y_{gt}^{'}=\left \{ y_{row}, y_{col} \right \} $ is the ground truth before and after flipping, $y_{row}$ and $y_{col}$ represent the ground truth after row and column transformation, $\lambda_{1}$ and $\lambda_{2}$ represent the weighting coefficient, $M$ represents the number of categories and $N$ is batch size.
To balance the speed and efficiency of the model under a limited computational budget, we adopt different moving steps and thresholds to improve the representation ability of the network and reduce the computational complexity. 

\begin{table*}[t]
\caption{Individual category results and mIoU scores on the CityScapes dataset.}
\begin{center}
\setlength{\tabcolsep}{0.65mm}{
\begin{tabular}{c|ccccccccccccccccccc|c}
\hline
\textbf{Model} & \textbf{Roa} & \textbf{Sid} & \textbf{Bui} & \textbf{Wal} & \textbf{Fen} & \textbf{Pol} & \textbf{TLi} & \textbf{TSi} & \textbf{Veg} & \textbf{Ter} & \textbf{Sky} & \textbf{Ped} & \textbf{Rid} & \textbf{Car} & \textbf{Tru} & \textbf{Bus} & \textbf{Tra} & \textbf{Mot} & \textbf{Bic} & \textbf{mIoU} \\
\hline
SegNet \cite{Segnet} & 96.4 & 73.2 & 84.0 & 28.4 & 29.0 & 35.7 & 39.8 & 45.1 & 87.0 & 63.8 & 91.8 & 62.8 & 42.8 & 89.3 & 38.1 & 43.1 & 44.1 & 35.8 & 51.9 & 57.0 \\
ENet \cite{ENet} & 96.3 & 74.2 & 75.0 & 32.2 & 33.2 & 43.4 & 34.1 & 44.0 & 88.6 & 61.4 & 90.6 & 65.5 & 38.4 & 90.6 & 36.9 & 50.5 & 48.1 & 38.8 & 55.4 & 58.3 \\
ICNet \cite{ICNet} & 97.1 & 79.2 & 89.7 & 43.2 & 48.9 & 61.5 & 60.4 & 63.4 & 91.5 & 68.3 & 93.5 & 74.6 & 56.1 & 92.6 & 51.3 & 72.7 & 51.3 & 53.6 & 70.5 & 69.5 \\
LEDNet \cite{LEDNet} & 98.1 & 79.5 & 91.6 & 47.7 & 49.9 & 62.8 & 61.3 & 72.8 & 92.6 & 61.2 & 94.9 & 76.2 & 53.7 & 90.9 & 64.4 & 64.0 & 52.7 & 44.4 & 71.6 & 70.6 \\
DDRNet-s \cite{DDRNet} & 98.1 & 84.4 & 92.0 & 53.3 & 59.4 & 61.2 & 68.8 & 76.3 & 92.1 & 65.2 & 94.1 & 80.1 & 60.9 & 94.8 & 84.3 & 88.3 & 76.8 & 61.2 & 75.4 & 77.2 \\
\hline
BCMFNet (Ours) & \textbf{98.2} & \textbf{84.6} & \textbf{92.2} & \textbf{54.0} & \textbf{61.3} & \textbf{64.8} & \textbf{71.4} & \textbf{78.1} & \textbf{92.7} & \textbf{69.8} & \textbf{94.9} & \textbf{81.7} & \textbf{64.0} & \textbf{95.1} & 81.4 & 87.9 & \textbf{79.0} & 60.0 & \textbf{75.8} & \textbf{78.2} \\
\hline
\end{tabular}}
\end{center}
\vspace{-15pt}
\label{ind}
\end{table*}

\vspace{-8pt}
\section{Experiments}
\subsection{Datasets and Metrics}
Our method is evaluated on the public Cityscapes \cite{Cityscapes} and CamVid \cite{CamVid} datasets.
Cityscapes is a dataset of urban landscapes, which contains a total of 5000 images. 2975 images are used for the training set, 500 images are used for the validation set, and 1525 images are used for the test set. The dataset includes 19 categories, and the resolution of the images is 2048×1024.
CamVid contains a total of 701 images, with 367 images for the training set, 101 images for the validation set, and 233 images for the test set. There are 32 categories in total. 11 categories are used for the semantic segmentation task, and the resolution of the images is 960×720.\par
In experiments, we employ mean cross-union (mIoU) to evaluate model accuracy and use FPS, GFLOPs, and Params to measure model efficiency.\par

\begin{table}[t]
\vspace{-5pt}
\caption{Accuracy and speed comparison on Cityscapes dataset.}
\begin{center}
\setlength{\tabcolsep}{1.9mm}{
\begin{tabular}{c|cccc}
\hline
\textbf{Model} & \textbf{mIoU} & \textbf{FPS} & \textbf{GFLOPs} & \textbf{Params} \\
\hline
SegNet\cite{Segnet} & 57 & 16.7 & 286 & 29.5M \\
ENet\cite{ENet} & 57 & 135.4 & 3.8 & 0.4M \\
BiSeNet1\cite{BiSeNet} & 68.4 & 105.8 & 14.8 & 5.8M \\
BiSeNetV2\cite{BiSeNetv2} & 72.6 & 156 & 21.1 & - \\
ICNet\cite{ICNet} & 69.5 & 30.0 & - & 7.8M \\
LEDNet\cite{LEDNet} & 70.6 & 71.0 & - & 0.94M \\
CABiNet\cite{CABiNet} & 75.9 & 76.5 & 12.0 & 2.64M \\
DDRNet-s\cite{DDRNet} & 77.4 & 101.6 & 36.3 & 5.7M \\
\hline
BCMFNet (Ours) & \textbf{78.2} & 103.4 & 35.8 & 5.6M \\
\hline
\end{tabular}}
\end{center}
\vspace{-1em}
\label{acc_Cityscapes}
\end{table}

\begin{table}[t]
\vspace{-5pt}
\caption{Accuracy and speed comparison on CamVid dataset.}
\begin{center}
\setlength{\tabcolsep}{7.7mm}{
\begin{tabular}{c|cc}
\hline
\textbf{Model} & \textbf{mIoU} & \textbf{FPS} \\
\hline
DFANet\cite{DFANet} & 64.7 & 120 \\
BiSeNet1\cite{BiSeNet} & 65.6 & 175 \\
BiSeNetV2\cite{BiSeNetv2} & 72.4 & 124 \\
MSFNet\cite{MSFNet} & 75.4 & 91 \\
DDRNet-s\cite{DDRNet} & 74.7 & 230 \\
\hline
BCMFNet (Ours) & \textbf{76.2} & 225 \\
\hline
\end{tabular}}
\end{center}
\vspace{-16pt}
\label{acc_CamVid}
\end{table}

\begin{table}[t]
\vspace{-5pt}
\caption{Comparison results of the proposed modules on Cityscapes dataset.}
\begin{center}
\setlength{\tabcolsep}{4.1mm}{
\begin{tabular}{c|ccc|c}
\hline
\textbf{Index} & \textbf{BASE} & \textbf{LMFM} & \textbf{BCL} & \textbf{mIoU} \\
\hline
1 & \Checkmark & \XSolidBrush & \XSolidBrush & 76.8 \\
2 & \Checkmark & \Checkmark & \XSolidBrush & 77.6 \\
3 & \Checkmark & \Checkmark & \Checkmark & \textbf{78.2} \\
\hline
\end{tabular}}
\end{center}
\vspace{-16pt}
\label{ablation}
\end{table}

\vspace{-6pt}
\subsection{Implementation Details}
\vspace{-3pt}

The structure of the model is described in Section \ref{method}.
For Cityscapes dataset, we randomly crop the input size to 1024×1024. We use the SGD optimizer with the momentum of 0.9 and the weight decay of 0.0005. We train models with a batch size of 8 and a initial learning rate of 0.1.
For CamVid dataset, we randomly crop the input size to 960 × 720, set the batch size to 16, and the rest of the settings are the same as Cityscapes dataset.\par 

\vspace{-6pt}
\subsection{Results}
\vspace{-3pt}

In experiments, we compare our method with state-of-the-art real-time semantic segmentation methods. Results on Cityscapes dataset are shown in Table \ref{acc_Cityscapes}. Our method achieves the best balance between real-time speed and high accuracy and is faster than baseline for inference. As can be seen from the single category results in the Table \ref{ind}, out of 19 categories, we have 16 the best scores. It can be seen from the Table \ref{ind} and Figure \ref{results} that our method has higher accuracy for categories with large intersection areas and well-defined boundaries, such as roads, fences, traffic lights, etc. Results on CamVid dataset are shown in Table \ref{acc_CamVid}. Our results show a 1.5\% improvement over baseline while maintaining comparable inference speed. Figure \ref{results} shows partial results on the CamVid dataset.

\vspace{-6pt}
\subsection{Ablation Study}
\vspace{-3pt}

In this section, we further discuss and analyze the effectiveness of key components in BCMFNet.
We conduct experiments on the difficult dataset under the fixed backbone network by controlling other conditions. As shown in Table \ref{ablation}, BASE(Index-1) is without our proposed LMFM and BCL components. Compared to baseline, LMFM brings 0.8\% improvement, and BCL brings 0.6\% improvement. Using LMFM and BCL at the same time achieves the best performance, with the accuracy increasing from 76.8\% to 78.2\% with little impact on the model's inference speed. The results show the good complementarity of LMFM and BCL, which can jointly optimize the model.

\vspace{-4pt}
\section{Conclusion}
\vspace{-1pt}
In this paper, we are devoted to the real-time semantic segmentation of road scenes. In our BCMFNet, a new module LMFM is introduced, which connects the low-resolution multi-scale feature maps of the model to obtain richer contextual information while reducing the amount of computation. 
In addition, we propose the BCL to enhance learning of hard samples by detecting boundary-hard samples online. 
The experimental results show that the proposed method has better performance under the same conditions compared with the existing models.

\clearpage
\bibliographystyle{IEEEbib}
\begin{spacing}{1.0}
\bibliography{refs}
\end{spacing}

\end{document}